\documentclass[acmtog]{acmart}
\renewcommand\footnotetextcopyrightpermission[1]{}
\usepackage{booktabs} % For formal tables
\usepackage{cleveref}
\usepackage{graphicx}
\usepackage{enumitem}
\usepackage{siunitx}
\usepackage{subcaption}
\usepackage{multirow}
\usepackage{wrapfig}
\usepackage{tabularx}
\usepackage{natbib}
\usepackage{animate}
\usepackage{ocgx2}

\newcommand{\ra}[1]{\renewcommand{\arraystretch}{#1}}
\newcommand{\gbox}[1]{\colorbox[HTML]{CCFFC4}{\makebox[1.2em]{\rule[1pt]{0pt}{5pt}#1}}}
\newcommand{\rbox}[1]{\colorbox[HTML]{FECDCD}{\makebox[1.2em]{\rule[1pt]{0pt}{5pt}#1}}}

\usepackage[ruled]{algorithm2e} % For algorithms

\SetAlFnt{\small}
\SetAlCapFnt{\small}
\SetAlCapNameFnt{\small}
\SetAlCapHSkip{0pt}

\begin{document}
% Title portion
\title{Towards Spatially-Varying Gain and Binning}

% DO NOT ENTER AUTHOR INFORMATION FOR ANONYMOUS TECHNICAL PAPER SUBMISSIONS TO SIGGRAPH 2019!
\author{Anqi Yang}
\affiliation{%
 \institution{Carnegie Mellon University}
 \streetaddress{5000 Forbes Avenue}
 \city{Pittsburgh}
 \state{PA}
 \postcode{15213}
 \country{USA}}
\email{anqiy1@andrew.cmu.edu}
\author{Eunhee Kang}
\affiliation{%
 \institution{Samsung Advanced Institute of Technology}
 \city{Yeongtong-gu, Suwon-si, Gyeonggi-do}
 \country{South Korea}
}
\email{eunhee.kang@samsung.com}
\author{Wei Chen}
\affiliation{%
 \institution{Carnegie Mellon University}
 \streetaddress{5000 Forbes Avenue}
 \city{Pittsburgh}
 \state{PA}
 \postcode{15213}
 \country{USA}}
 \email{wc3@andrew.cmu.edu}
\author{Hyong-Euk Lee}
\affiliation{%
 \institution{Samsung Advanced Institute of Technology}
 \city{Yeongtong-gu, Suwon-si, Gyeonggi-do}
 \country{South Korea}
}
\email{hyongeuk.lee@samsung.com}
\author{Aswin C. Sankaranarayanan}
\affiliation{%
 \institution{Carnegie Mellon University}
 \streetaddress{5000 Forbes Avenue}
 \city{Pittsburgh}
 \state{PA}
 \postcode{15213}
 \country{USA}}
 \email{	saswin@andrew.cmu.edu}

\renewcommand\shortauthors{Yang, A. et al}

\begin{abstract}
Pixels in image sensors have progressively become smaller, driven by the goal of producing higher-resolution imagery.
However, \textit{ceteris paribus}, a smaller pixel accumulates  less light, making image quality worse.
This interplay of  resolution, noise and the dynamic range of the sensor and their  impact on the eventual quality of acquired imagery is a fundamental concept in photography.
In this paper, we propose spatially-varying gain and binning to enhance the noise performance and dynamic range of image sensors.
First, we show that by  varying  gain \textit{spatially} to local scene brightness, the read noise can  be made negligible, and the dynamic range of a sensor is expanded by an order of magnitude.
Second, we propose a simple analysis to find a binning size that best balances resolution and noise for a given light level; this analysis predicts a spatially-varying  binning strategy, again based on local scene brightness, to effectively increase the overall signal-to-noise ratio. % without sacrificing resolution.
We discuss analog and digital binning modes and, perhaps surprisingly, show that digital binning outperforms its analog counterparts when a larger gain is allowed.
Finally, we demonstrate that combining spatially-varying gain and binning  in various applications, including high dynamic range imaging, vignetting, and lens distortion.
\end{abstract}



\maketitle
\pagestyle{empty}
\thispagestyle{empty}

\section{Introduction}
\label{sec:intro}
\begin{figure}
	\centering
	\includegraphics[width=1.0\linewidth]{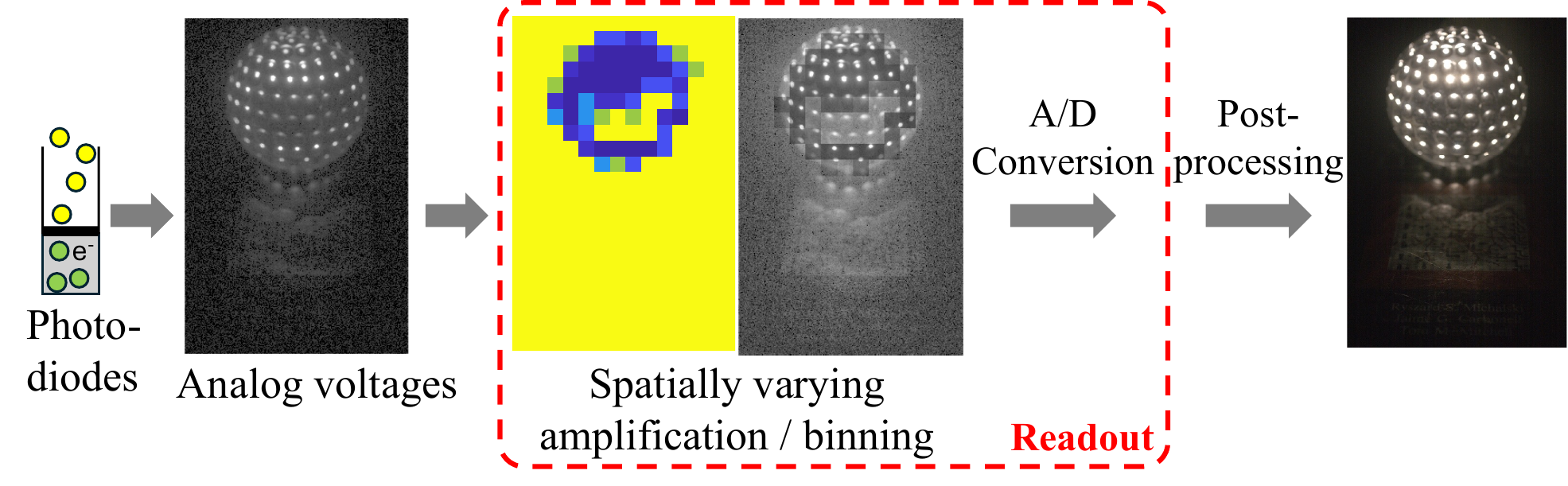}
%	\animategraphics[loop, autoplay, width=1.0\linewidth]{0.3}{figures/teaser/teaser_}{006}{009}
	 \makebox[0pt][l]{%
		\begin{ocg}[radiobtngrp=letters]{A}{A}{on}
			\includegraphics[width=\linewidth]{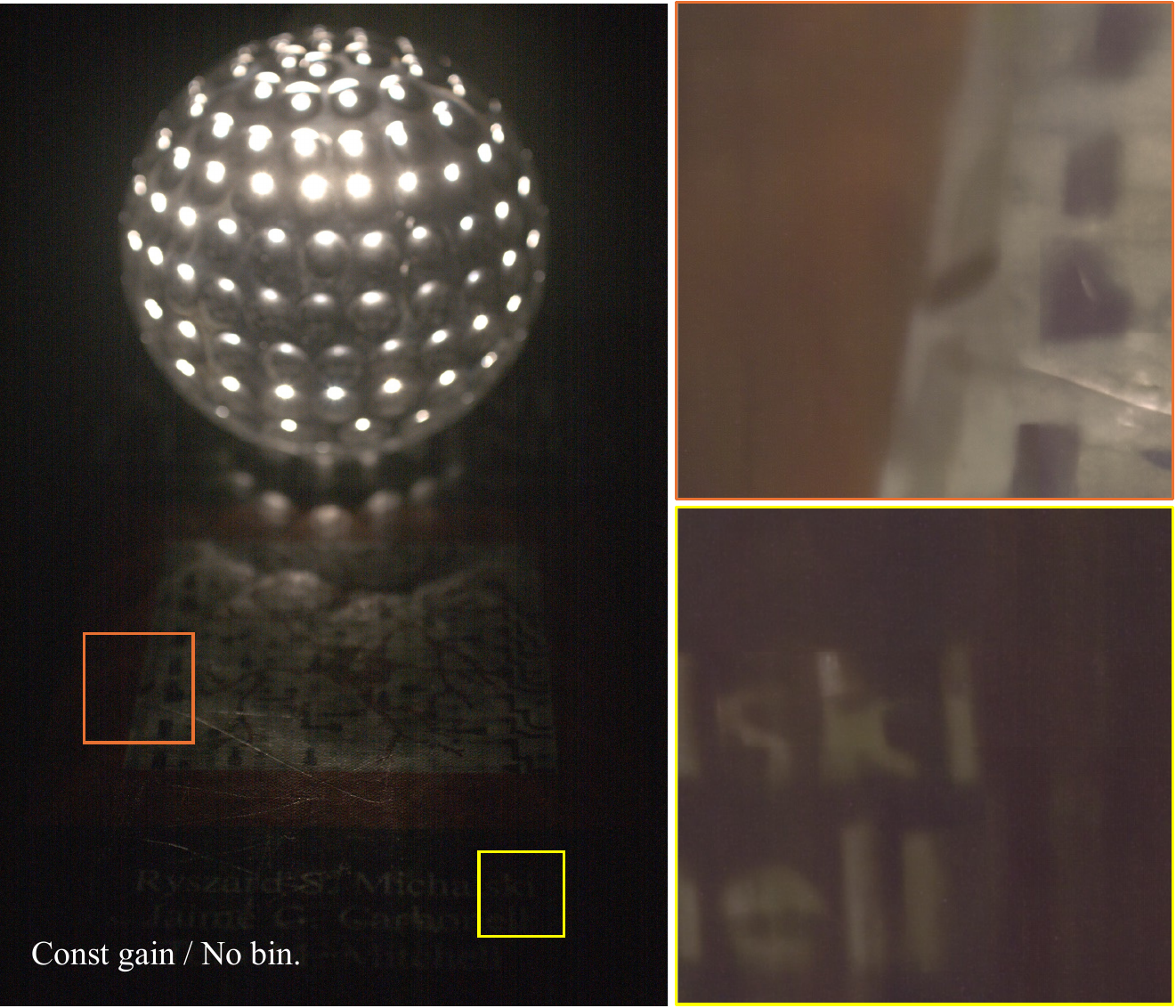}
		\end{ocg}
	}%
	\makebox[0pt][l]{%
		\begin{ocg}[radiobtngrp=letters]{B}{B}{off}
			\includegraphics[width=\linewidth]{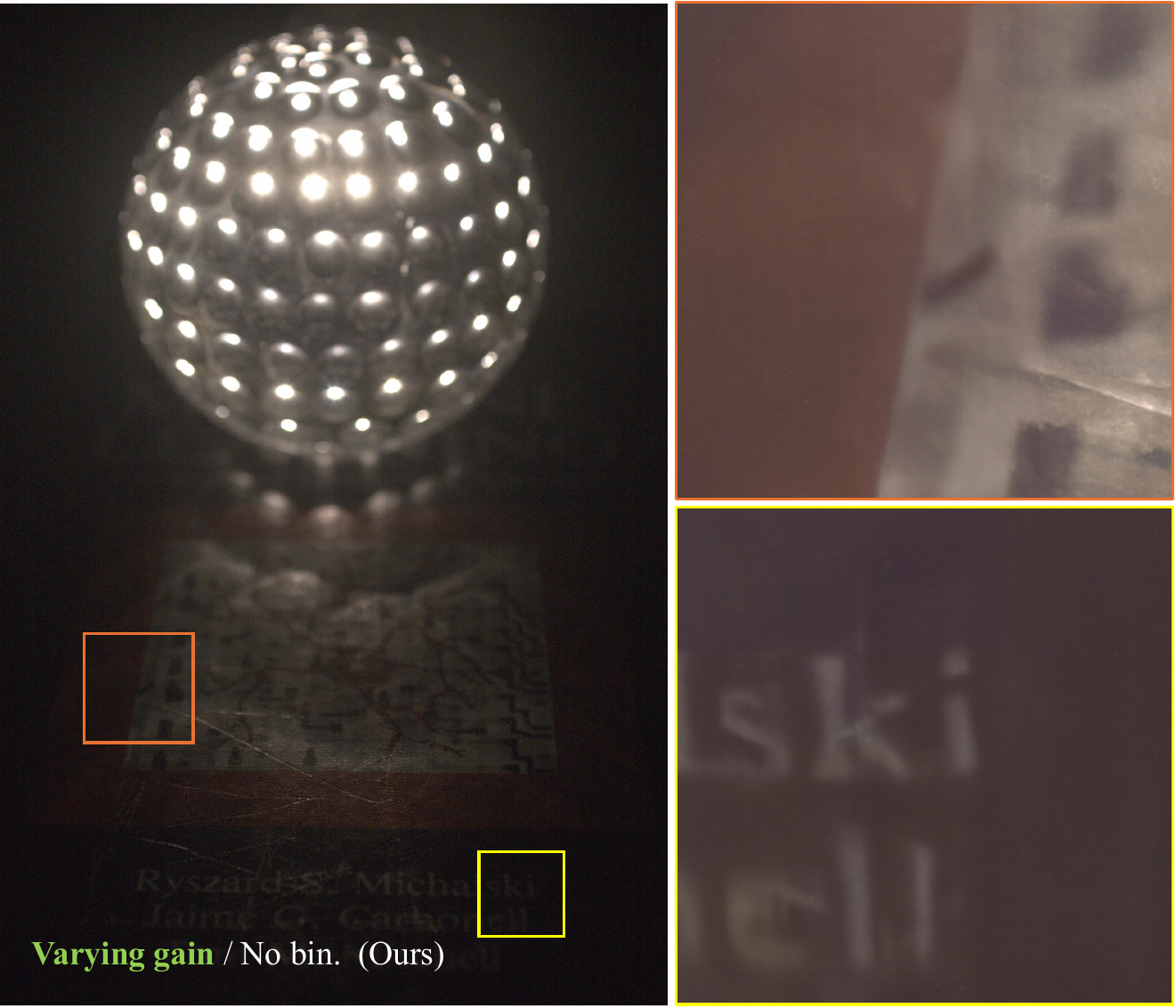}
		\end{ocg}
	}%
	\begin{ocg}[radiobtngrp=letters]{C}{C}{off}
		\includegraphics[width=\linewidth]{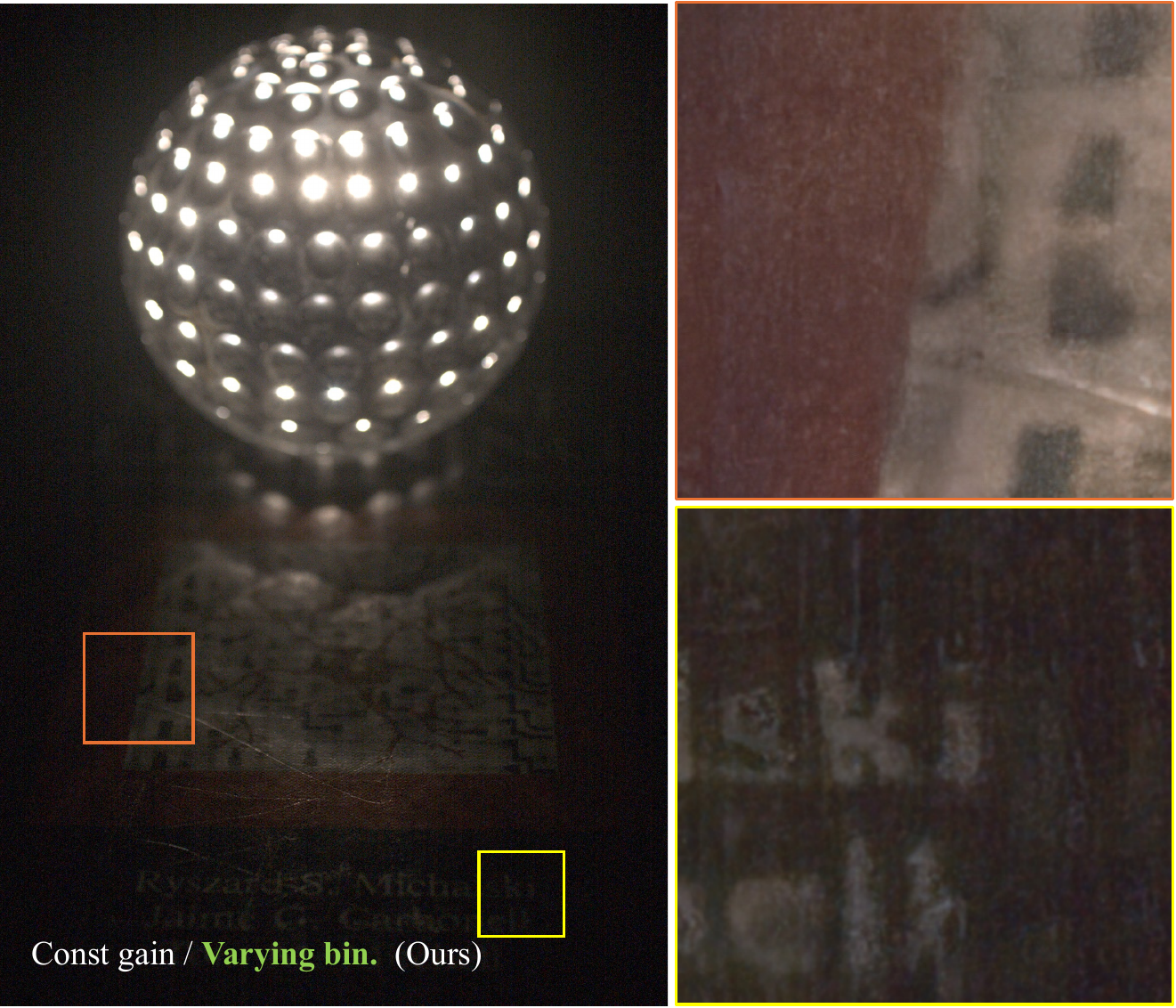}
	\end{ocg}
	
\showocg{A}{\fcolorbox{black}{gray!40}{\parbox[][8pt]{1.8cm}{} \footnotesize \textbf{Conventional}}}\quad
	\showocg{B}{\fcolorbox{black}{gray!40}{\parbox[][8pt]{2.5cm}{} \footnotesize \textbf{Proposed vary. gain}}}\quad
	\showocg{C}{\fcolorbox{black}{gray!40}{\parbox[][8pt]{2.5cm}{} \footnotesize \textbf{Proposed vary. binning}}}
	\caption{\textbf{Proposed spatially-varying readout techniques.} 
	The upper figure is an illustration of the proposed spatially-varying readout techniques. The lower figures are captured by BFS-U3-200S6C machine vision camera and denoised by SOTA transformer-based method Restormer~\cite{zamir2022restormer}. \textbf{Note: We kindly request readers to use Adobe Acrobat Reader to interact with the clickable buttons.} Conventional sensor uses a constant gain and no binning. Clicking between conventional and the proposed spatially-varying gain, proposed readout strategy produces much more details. Clicking between conventional and the proposed spatially-varying binning, proposed retains better contrast due to higher signal-to-noise ratio.}
	\label{fig:new-teaser}
\end{figure}

Noise and resolution are central to an image sensor, affecting the quality of the photographs acquired by it and the flavor of algorithmic post-processing required.
The importance of these two factors is readily seen across a wide game: from classic problems such as denoising and super-resolution to more modern ones pertaining to (high) dynamic range.
All of these challenges are routinely encountered and addressed, to some extent, \textit{every time} a photograph is acquired.
Hence, advancing the design of image sensors---the premise of this work---to combat noise and resolution can have an outsized impact on photography as well as the myriad set of applications that benefit from visual imagery.

At capture time, a sensor and its associated electronics do offer choices to a photographer to control noise and resolution; this comes in the form of gain (or ISO) and binning.
Gain refers to a pre-amplification of the signal before readout. Using a high gain, for example, to amplify a weak signal before digitization helps in suppressing the effects of quantization.
However, a large gain also suppresses read noise, a dominant source of noise that is caused by electronics in the sensor.
This  improves quality in the dark regions as they are read noise dominated. 
However, maxing out the gain for dark regions would end up saturating the bright regions in the same scene, limiting the use of an extremely large gain. 
Binning, on the other hand, involves adding the charge at neighboring pixels to increase signal levels.
Photon noise depends on light levels, and increasing light levels by binning increases the signal-to-noise ratio (SNR).
However, binning produces larger pixels; applying the same binning size to the entire sensor would unnecessarily sacrifice fine details in the bright regions that are already resolved in high SNRs.

This paper makes the argument for novel capabilities in image sensors in the form of \textit{spatially-varying and scene adaptive} gain and binning.
At its simplest incarnation, imagine if we had a sensor which at readout allows for each patch to be readout with a different gain and binning.
The argument for a spatially-varying gain is immediately evident since, for each patch, we can select the largest gain that avoids saturation for the pixels within.
Since dynamic range observed in a patch is bound to be significantly smaller than that observed in the entire image, the darkest patches will benefit from using the highest gain offered by the imager without risking saturation at brighter regions.
Effectively, this expands the dynamic range of the sensor by reducing the noise floor.
However, this will require some knowledge of the bright and dark patches in the scene, which we can obtain from a low-resolution snapshot.
We also discuss a single-shot variant that implements a per-pixel spatially-varying gain, using the intensity observed at a previously readout pixel.

Spatially-varying binning poses a different question: can binning, which is explicitly a loss of resolution, ever improve the quality  of the acquired photograph?
To answer this question, we  develop a simple theory that, given the light level of the scene, analyzes the optimal binning size that resolves  features in the photograph. %at a certain SNR threshold.  
Surprisingly, a larger binning in  dark regions gives better resolution, 
since our ability to resolve details is also strongly dependent on  noise~\cite{treibitz2012resolution}.
We also analyze three binning modes: analog additive, analog average, and digital binning. We also show that digital binning  can achieve better performance than both analog binning modes, when a larger gain is allowed for.

Finally, we combine spatially-varying gain and binning to reduce both read and photon noise for high dynamic range imaging, vignetting, and spatially-varying lens distortion. 
Figure \ref{fig:new-teaser} shows an example of the benefits to be derived using our proposed techniques.

\paragraph{Contributions.} This paper revisits concepts of noise, resolution, and dynamic range for image sensors, through the mindset of rethinking gain and binning.
\begin{itemize}[leftmargin=*]
	\item \textit{Spatially-varying gain.} We propose spatially-varying gain that adapts to local scene brightness, which significantly reduces read noise for dark regions  and expands sensor dynamic range.
	\item \textit{Spatially-varying binning.} We establish an  analysis that maps light levels to  optimal binning sizes and apply it to spatially-varying binning, thereby achieving better noise-resolution tradeoffs.
	\item \textit{Applications.} Proposed techniques show significant improvement in noise performance for  high dynamic range imaging, vignetting, and lens distortion.
\end{itemize}

\paragraph{Limitations.} While the proposed ROI-based techniques are relatively straight-forward to implement,  per-pixel varying gain requires a modification to the  readout circuits, capabilities that we are yet to implement in hardware.
There is an inherent risk: changing readout circuitry could increase read noise, which might annul the improvements in the noise performance predicted by these emulations.

\paragraph{Impact.} Our work  looks into addressing the conflict between resolution, noise, and dynamic range for sensors. 
While increasing dynamic range for brighter parts of the scene has been studied extensively, there are few techniques that address noise floor, the limiting factor for darker regions.
We show that applying spatially-varying gain, the sensor dynamic range can be expanded by an  order of magnitude and resolving more signals towards the low light end. 
Our work also provides a way to select binning, to tradeoff resolution and noise, for general photography.

\section{Related Work}
\label{sec:related}
This work touches upon noise and resolution which has been studied extensively in imaging and vision. 
%
%We briefly discuss some of the key related ideas.

\paragraph{Noise Analysis.} Early work including ~\citet{clarkreview} and
\citet{healey1994radiometric} look at models for understanding noise in image sensors.
%
%Noise in image sensors come in various forms spanning photon noise, read noise, dark current, fixed pattern noise and dead pixels.
%
Photon noise is caused due to randomness in photon arrivals at the sensor, and can be modeled as being Poisson distributed. 
%
%It has an SNR equal to the square root of the average number of photons arriving at the pixel for the exposure duration used.
%
Read noise, caused by voltage fluctuations in readout circuitry, is introduced at both pre- and post-amplifier stages. 
\citet{hasinoff2010noise} provide a detailed model of read noise and the role of sensor gain in the context of high dynamic range (HDR) photography.
They, and others  \cite{martinec2008},  point out how pre-amplifier read noise is often significantly smaller than its post-amplifier counterpart .
This suggests using a larger gain to significantly suppress post-amplifier read noise.
Noise and dynamic range are intricately coupled.
However, the majority of computational cameras devoted to HDR  imaging  \cite{narasimhan2005enhancing,sun2020learning,nayar2004programmable} focus on enhancing range at the brighter end of the light levels; this is relatively easier as it involves blocking light.
A notable exception is a recent sensor \cite{imx} that uses microlensets of different sizes to redistribute light, providing an assorted pixel-like design, that does increase light levels at some pixels without using increased exposure times.
Outside of this, there are few techniques that suppress the noise floor to enhance darker regions of the photograph---the premise of this work.

\paragraph{Spatially-varying gain.}
\citet{hajsharif2014hdr} propose a sensor where the gain is varied across pixels, with a spatial tiling that is pre-determined; for example, alternative rows of the sensor have gains of $1\times$ or  $16\times$, respectively.
However, the gain pattern is fixed, and not adaptive to the specifics of the scene. 
To adapt the multiplexing patterns to different scenes, \citet{qu2024spatially} propose an enumeration method that selects the best gain and exposure pattern according to a pilot shot of the scene.
Although scene adaptive, this method still uses a global multiplexing pattern for the entire image, resulting in an inherent loss of resolution at the brightest and darkest pixels.
In contrast, our approach aims to avoids the loss of resolution by applying locally varying gain and binning patterns.

\paragraph{Pixel binning.}  %Prior works look at the design of the binning pattern and restoration algorithms~\cite{zhang2018pixel,jin2012analysis}.
\citet{zhang2018pixel} propose a new pixel binning pattern for color sensors that minimizes the binning artifacts. An extension pattern design equals to a sensor interlacing original pixels with super-pixels binned from four neighbouring pixels. 
%The design is shown to be effective for single-shot HDR imaging.
%
\citet{jin2012analysis} analyze the analog additive binning on color sensors and design a specialized demosaic algorithm to suppress the binning artifacts. 
However, these techniques cannot adaptive to the local scene brightness.

\paragraph{Noise and resolution.} The idea that noise influences resolution has been studied formally in prior work.
\citet{treibitz2012resolution} look at this interplay  in the context of imaging in fog, showing how resolution loss happens not just due to loss of contrast in fog, but also due to  sensor noise.
We borrow the same formalism, but instead look at pixellation in place of fog.
This requires certain modifications to the underlying theory, based on frequency domain methods, where the effect of pixellation and noise are readily understood.

\paragraph{Computational sensors.} 
Focal plane sensor processors~\cite{nguyen2022learning,zarandy2011focal,carey2013mixed} have become more prevalent recently. Each pixel, containing a processing element next to the photodiode, can take in digital instructions and carry out on-chip analog and digital computation. However,  due to the additional circuits at each photosite, these sensors have large pixel pitches and are typically of low resolution and not suitable for high-quality photography.  
Another type of computational sensor is a programmable coded exposure sensor~\cite{luo2017exposure,sarhangnejad20195},  which adapts the pixelwise exposure time to input control signal. \citet{ke2019extending} realize scene-adaptive coded exposure by generating exposure codes from the readout of the previous frame. To improve the SNR of the dark regions, these sensors require a longer exposure time. 
%In contrast, ~\citet{hasinoff2010noise} shows using a high ISO benefits worst case SNR and operates with a shorter capture time. 
In contrast, our techniques fall into the category of programming gain and binning to increase noise performance, without the need of using a longer exposure time, and only involve modification to the sensor readout.

Another class of techniques work on innovating the analog to digital conversion. For example, \citet{gulve2023dual} propose a regression-based Flux-to-Digital Conversion in place of conventional analog-to-digital conversion, a new readout strategy that occurs concurrently with exposure and avoids saturating high flux, thereby extending the dynamic range of the sensor. These works provide alternative approaches to the proposed innovation in this paper.

\section{Noise in Image Sensors}
\label{sec:noise}

%\begin{figure}[!tt]
%	\centering
%	\includegraphics[width=\linewidth]{figures/sensor-pipeline.pdf}
%	\caption{A diagram illustrating CMOS sensor converts incident photons to digital values.}
%	\label{fig:sensor-pipeline}
%\end{figure}

Noise in image sensors mainly consists of three types: photon noise, read noise, and dark current.
A measured image $i$ can be formed as: %(see \cref{fig:sensor-pipeline}):
\begin{equation}
	i = \Phi \{g \cdot (l + n_{D} + n_{pre} ) + n_{post} \} + i_0,  \quad  l \sim Poisson(l^*). \label{eq:noise-model}
\end{equation}
Here, $l$ is the measured photon counts and follows a Poisson distribution with mean and variance as the expected photon arrival within the exposure time, $l^*$. 
We also assume that the sensor has a quantum efficiency of one; alternatively, we can replace the average photon arrivals with the average photo-electron arrivals, and absorb the quantum efficiency into $l^*$. 
The term $n_D$ denotes the dark current and scales linearly with exposure time and temperature. Since we discuss photography with an exposure time of up to hundreds of milliseconds, the dark current is negligible. Finally, $n_{pre}$ and $n_{post}$ are pre- and post-amplifier read noise. Both are signal-independent and follow Gaussian distributions with a mean of zero and variance of $\sigma_{pre}$ and $\sigma_{post}$. Note that for most sensors, $\sigma_{post}$ is one or two magnitudes larger than $\sigma_{pre}$, and thus post-amplifier read noise is much more significant than pre-amplifier read noise. $g$ is the analog gain. It usually ranges from one to hundreds. $\Phi$ denotes the analog-to-digital conversion (ADC). Since most sensors have a bit depth higher than their dynamic range, we can safely assume that the ADC noise is small. $i_0$ is the black level.

With the abovementioned assumption, we can simplify the noise model and estimate photon counts from the noisy digital image,
\begin{equation}
	\hat{l} = \Phi^{-1}(i-i_0)/g = l  + n_{pre} + n_{post} / g.
\end{equation}
$\smash{\hat{l}}$ has an expectation of $l^*$ and a total variance of $\smash{l^* + \sigma^2_{pre} + \sigma^2_{post} / g^2}$.
And SNR equals to  $\smash{l^* / \sqrt{l^* + \sigma^2_{pre} + \sigma^2_{post} / g^2}}$.

We can draw two key insights from the noise model.
\begin{itemize}[leftmargin=*]
	\item \textit{Constant gain.} A large gain $g$ can largely reduce the post-amplifier read noise term $\smash{\sigma^2_{post} / g^2}$. Since $\smash{\sigma^2_{pre}}$ is much smaller than $\sigma^2_{post}$, both read noises can be suppressed.
 This largely benefits low-light photography where read noise standard deviation is at a similar scale as the signal $l^*$. 
As a consequence, photon noise becomes the dominant source of noise for all practical light levels.
%  Large gain overcomes the effect of read noise and increases SNR to approximately $\sqrt{l^*}$, which is only photon noise limited. 
However, for a scene with a large dynamic range, increasing gain would saturate the bright regions, which limits the choice of a large gain.

\item \textit{Constant binning.} Binning neighboring pixels together increases SNR.  Averaging $N$ pixels results in an expectation of $l^*$ and a standard deviation of $\smash{\sqrt{l^* / N}}$, considering $\smash{l^* \gg \sigma^2_{pre} + \sigma^2_{post}/g^2}$. Thus the overall SNR is increased from $\smash{\sqrt{l^*}}$ to  $\smash{\sqrt{N l^*}}$. However, binning comes with a side effect of pixelation and loss of resolution. The binning size $N$ that produces the highest SNR for the low-light regions could sacrifice fine details in the bright regions. 
\end{itemize}

\paragraph{Overview.} We propose spatially-varying gain and binning to overcome read- and photon-noise limitations, and expand the dynamic range of a sensor.
 In \cref{sec:sv_gain}, we show that setting varying gain for varying signal levels in a single shot effectively reduces the read noise for dark scenes without saturating bright regions. In \cref{sec:sv_binning}, we propose a spatially-varying binning strategy, where pixel binning size is decided by the scene light level. 

\section{Spatially-Varying Gain}
\label{sec:sv_gain}
We propose to apply a spatially-varying gain, and discuss its ability to reduce read noise and expand  dynamic range, as well as approaches for implementation.

\begin{figure*}[t]
	\centering
	\includegraphics[width=\linewidth]{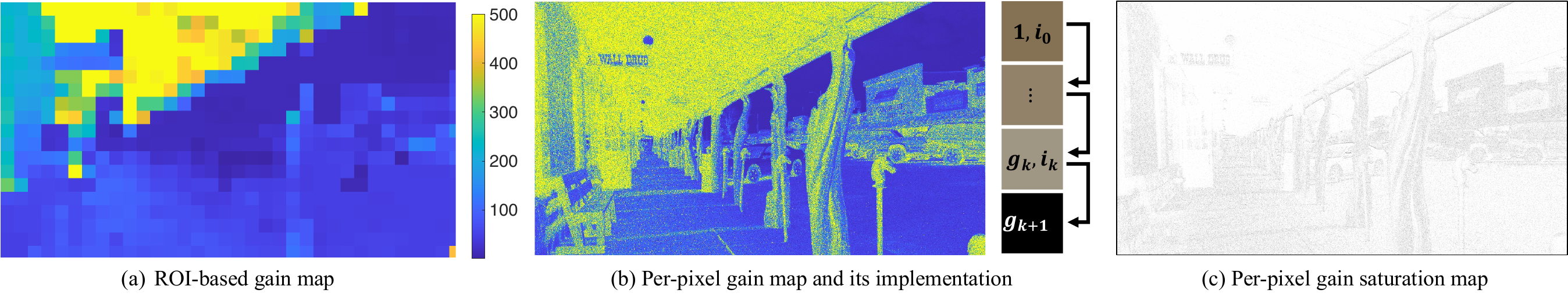}
	\caption{\textbf{Implementation of spatially-varying gain.} (a) ROI-based implementation first fragments the image into multiple ROIs and then sets a gain for each ROI based on snapshot light levels. (b) per-pixel implementation adaptively sets gain according to the readout of the previous pixel. Black dots in (c) show saturated pixels with per-pixel implementation. Only around \SI{3}{\percent} of pixel saturates.}
	\label{fig:implementation}
\end{figure*}

\paragraph{Choice of gain.}

Given an estimated scene light level $\smash{\hat{l}, \mathbb{E}(\hat{l}) = l^*}$, we aim to find the \textit{largest} gain such that the amplified signal $\smash{g l}$ saturates with a small probability.
We adopt the Gaussian-Heteroskedastic noise model~\cite{foi2008practical} and approximate the amplified signal with a Gaussian distribution,
$\smash{g \cdot(l + n_{pre})  + n_{post} \sim \mathcal{N} (g\hat{l}, g^2 \hat{l} +}$ $\smash{g^2\sigma^2_{pre} + \sigma^2_{post})}$.
Note that $g\hat{l}$ is intended to be close to well-capacity $l_{wc}$ and is much larger than the read noise variances, thus the total variance can be further simplified to $\smash{g^2 \hat{l}}$.
To make the probability of saturation small, we set a gain value $g$ that satisfied
\begin{equation}
l_{wc} \approx g\hat{l} + \eta g \sqrt{\hat{l}}.
\label{eq:compute-gain}
\end{equation}
When $\eta = 2$, the pixel saturates with a probability of  $2.2\%$.

\subsection{Design of spatially-varying gain}
We provide two strategies for design for spatially-varying gain.

\paragraph{Two-shot ROI-based strategy.} We first capture a noisy snapshot using constant gain. We fragment the snapshot into multiple region-of-interests (ROIs), so that each ROI has a smaller dynamic range, and compute the optimal gain for each ROI. %Regions that have no bright pixels  have a large gain applied to all pixels within it. 
The computed gain map is used to inform the subsequent capture and readout.
The advance of ROI-based methods is that it only requires a minimal change to today's readout circuitry in the form of being able to select multiple ROIs, instead of one.
However, it requires two images, leading to potential of motion-related artifacts (although their effects are not in the form of blur as we discuss later).
% and regions is under direct sunlight thus a small gain is applied. 
%Note that ROI-based implementation requires an additional snapshot of the scene to determine the light levels.
%, ROIs, and gain for each ROI.

%\setlength{\columnsep}{1em}
%\setlength{\intextsep}{0em}
%\begin{wrapfigure}[10]{r}{0.35\linewidth}
%\vspace{-5mm}
%	\centering
%	\includegraphics[width=0.8\linewidth]{figures/perpixel-gain.pdf}
%	\vspace{-0.9em}
%	\caption{\textbf{Readout with per-pixel varying gain.}}
%	% The previous gain and readout intensity decide the next gain.}
%	\label{fig:perpixel-gain}
%\end{wrapfigure}

\paragraph{Single-shot per-pixel strategy.} The per-pixel strategy sets gain adaptively during readout and only requires a \textit{single shot}. Since natural images are typically piecewise smooth, we assume the light level of one pixel is similar to its neighbors. Therefore, we use the readout value of one pixel to set the gain for the subsequent pixel (see \cref{fig:implementation}(b)). Specifically, from the readout value and gain of the $k$-th pixel, we estimate its light level $l_k$, and use $l_k$ as an approximate estimate for  the next pixel's light level, $\smash{\widehat{l}_{k+1} \approx l_k}$. By substituting  $\smash{\widehat{l}_{k+1} }$ into \cref{eq:compute-gain}, we obtain the gain $g_{k+1}$ for $(k+1)$-th pixel and set it in the ADC circuits for readout. %Figure~\ref{fig:implementation}(b) shows a per-pixel gain map for reading out an image of a high dynamic range scene.
When $k-$th pixel saturates, we reset $g_{k+1}=1$. Emperically, we set $\eta=4$ and only around \SI{3}{\percent} of pixels saturates in the captured image, as shown in Fig.~\ref{fig:implementation}(c).

\subsection{Improving dynamic range in a single-shot}
Spatially-varying gain expands the dynamic range of a sensor by one or two magnitudes within a single exposure. This is because the dynamic range of a sensor is typically decided by the ratio of its well-capacity and read noise floor, $\smash{l_{wc} / \sqrt{\sigma^2_{pre} + \sigma^2_{post}/g^2}}$. To capture a high dynamic range scene, a conventional sensor with a constant gain is limited to use a small $g$ to avoid saturating the bright objects, and the variance of post-amplifier read noise is much larger than that of pre-amplifier read noise, thus $\smash{\sigma^2_{post} / g^2 \gg \sigma^2_{pre}}$ and the dynamic range of a conventional sensor is approximately $l_{wc} / \sigma_{post}$.
In contrast, the proposed spatially-varying gain uses large gain $g$ to capture dark regions, effectively reducing post-amplifier read noise $\smash{\sigma^2_{post} /g^2}$ to negligible, $\smash{\sigma^2_{post} / g^2 \ll \sigma^2_{pre}}$, and the resulting dynamic range becomes $l_{wc} / \sigma_{pre}$.
%
%\begin{equation}
%	\frac{l_{wc}}{\sigma_{pre}} \gg \frac{l_{wc}}{\sigma_{post}}
%\end{equation}
%
Moreover, spatially-varying gain effectively reduces the read noise and leaves photon noise the bottleneck of image quality. Next, we will look into reducing photon and read noise with a proposed technique, spatially-varying binning.

\section{Spatially-Varying Binning}
\label{sec:sv_binning}
Small pixels gather less lights, and one way to increase light levels is to bin pixels.
This raises the following question: what is the optimal binning size that maximizes image quality?
We show this optimal binning is tightly coupled with scene light levels.
This is because effective resolution increases as pixel gets smaller, but decreases with increasing noise levels---a side-effect of small pixels---indicating a sweet spot that best trades off resolution and noise.

\subsection{Analysis of the optimal pixel pitch}
Given a scene light level, what is the optimal pixel size that achieves the highest effective resolution?
We define \textit{effective resolution} as the frequency whose ratio between the noise-free signal contrast and the measured noise standard deviation is greater than a predefined threshold $\text{SNR}_t$. The resolution with SNR equals $\text{SNR}_t$ is the highest effective resolution, and frequencies smaller than it are all effective.

We characterize scenes of various feature sizes by examining sinusoidal signals with varying frequencies.
Consider a camera with an ideal lens and a sensor with a pixel pitch \SI[parse-numbers=false]{p}{\micro\meter}. The measured signal can be modeled as a convolution between the signal and a box function induced by the pixel size.
%\begin{equation}
%	l^*(x,y,f_0; l_0, p) = \{l_0 \cos(2 \pi f_0 x) + l_0 /2 \} * b \biggl\{ \frac{x}{p}, \frac{y}{p}\biggr\}.
%\end{equation}
This models the expected noise-free measurement by incorporating the blurring effect of pixel pitch. The contrast of the noise-free measurement of a sinusoid with frequency $f_0$ and a  light level of $l_0$ photons per unit area is,
\begin{equation}
	c(f_0; l_0, p) = |\max l^* - \min l^*| = l_0 p^2 \frac{sin(\pi p f_0)}{\pi f_0}.
\end{equation}
We refer readers to the appendix for detailed derivation. We plug in the expected signal $l^*$ to the noise model shown in \cref{eq:noise-model} and obtain the noisy measurements.  The total noise variance is,
\begin{equation}
	\sigma(f_0; l_0, p) = \sqrt{\sigma^2_{pre} + \sigma^2_{post} / g^2 + l_0 p^2 / 2},
\end{equation}
where read noise is independent of pixel size, and shot noise variance increases proportionally to pixel area $p^2$. 
With a specified threshold $\text{SNR}_t$, we can find the cutoff frequency $f_\text{cutoff}$ and all features below $f_\text{cutoff}$ are considered resolved.
%\begin{equation}
%	f_\text{cutoff}(l_0, p)= \arg\min_{f_0} \Biggl\| \frac{c(f_0; l_0, p)}{\sigma(f_0; l_0, p)} - \text{SNR}_t \Biggr\|
%\end{equation}
%
\begin{figure*}[t]
	\centering
	\includegraphics[width=1.0\linewidth]{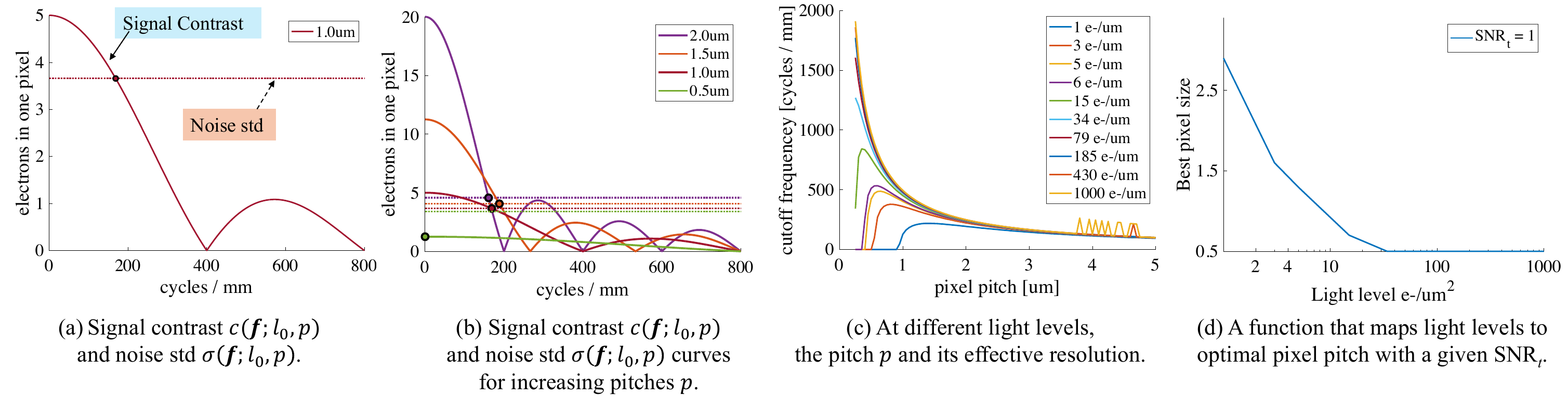}
	\caption{\textbf{Analysis of optimal binning.} Left figure shows the signal contrast, noise floor, and cutoff frequencies for pixel pitches from \SI{0.5}{\micro\meter} to \SI{2.0}{\micro\meter} under a fixed light condition. Right figure shows functions that map scene light levels to optimal pixel sizes under the required SNR threshold.}
	\label{fig:binning-theory}
\end{figure*}
%
%Fig.~\ref{fig:binning-theory}(a) shows one set of signal and noise curves for pitch $p$ and at a given light level $l_0$. Considering $\text{SNR}_t=1$, the two curves intersects at the cut-off frequency $f_\text{cutoff}(l_0, p)$. This means that given light level $l_0$, a sensor with pixel pitch $p$ can resolve features with maximum frequency $f_\text{cutoff}(l_0, p)$.
%As shown in \cref{fig:binning-theory}(b), we repeat this analysis varying pixel pitches, from \SI{0.5}{\micro\meter} to \SI{2.0}{\micro\meter}, given the same light level $l_0$, and find the pixel pitch $p^*$ that achieves the highest cutoff frequency, $p^*(l_0) = \arg\max_{p} f_\text{cutoff}(l_0, p)$. Note that the optimal pixel size is a function of light level, and decreases as the scene gets brighter, as shown in \cref{fig:binning-theory}(c). This confirms the intuition that small pixels suit bright scenes and large pixels suit dark scenes.
%The SNR threshold $\text{SNR}_t$ is a hyperparameter and is determined empirically by examining a set of measured image quality. Fig.~\ref{fig:binning-theory}(c) and (d) analyze with $\text{SNR}_t=1$ and $\text{SNR}_t=6$ respectively.

\begin{figure}[t]
	\includegraphics[width=1.0\linewidth]{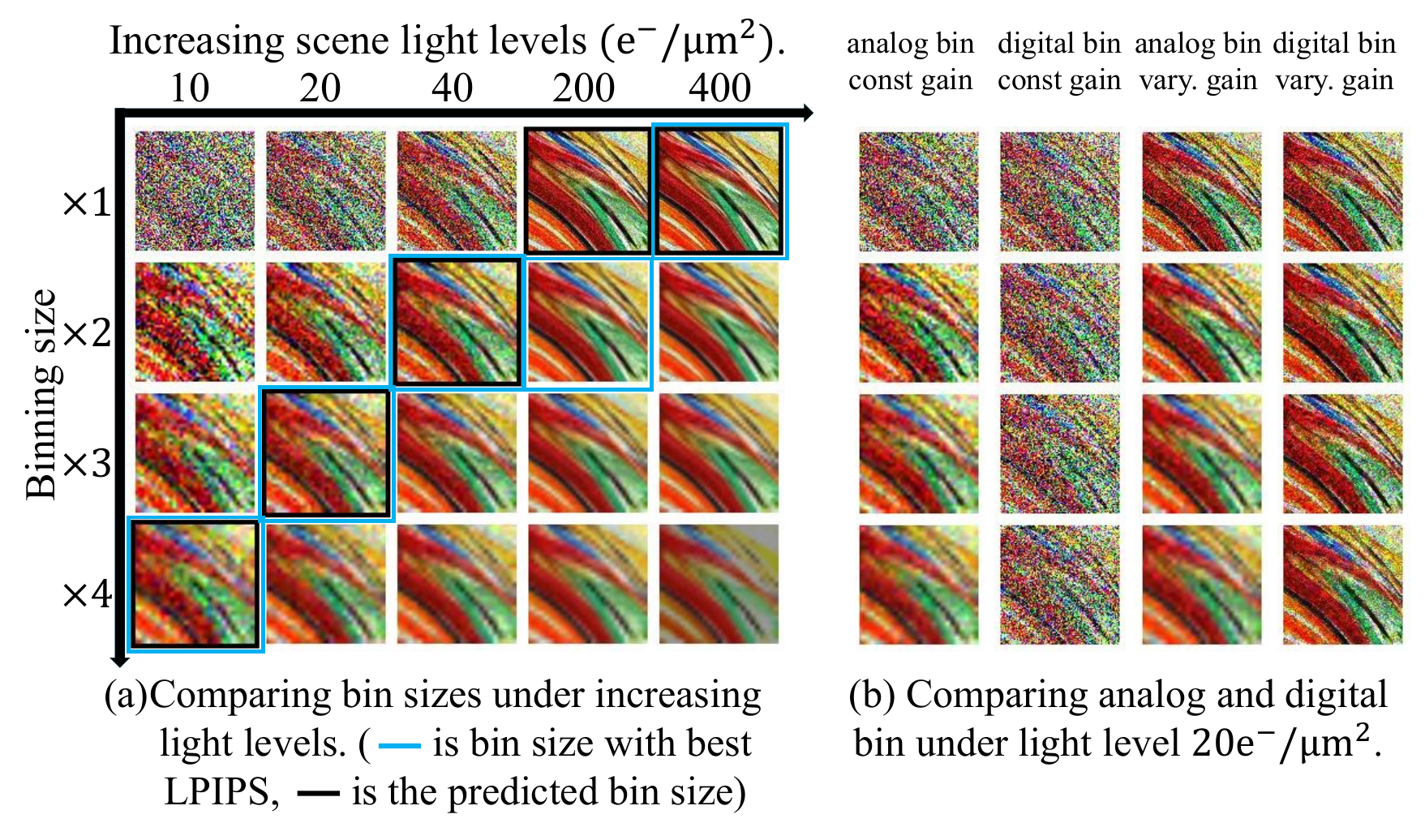}
	\caption{\textbf{Optimal bin sizes under different light levels.} (a) shows the effect of increasing bin sizes under from dim light condition to sufficient light. All are analog additive binning for unit pixel size \SI{0.5}{\micro\meter}. Black box shows the predicted optimal binning, and blue box shows the binning size with best LPIPS score~\cite{zhang2018unreasonable}. (b) compares analog and digital binning with small and large gain under dim light conditions.}
	\label{fig:sweep-bin}
\end{figure}

\begin{table}[h]	
	\footnotesize
	\centering
\ra{1.2}
	\caption{\textbf{A summary of binning modes.} Assume equal weights for neighboring pixels and normalize the combined signal to the same level.}	\label{tab:binning-modes}
	\begin{tabular}{@{}lll@{}} \toprule
Binning modes & Imaging model $\hat{l}$&Total noise variance\\ 		\midrule
		No binning   & $l + n_{pre}+ \frac{n_{post}}{g}$    & $l^* + \sigma^2_{pre} + \frac{\sigma^2_{post}}{g^2}$\\
		 Additive binning&   $\frac{1}{N}$$\sum_{i=1}^{N} \{l_i + n_{i,pre} \}+ \frac{n_{post}}{N\widehat{g}}$    & $\frac{l^*}{N} + \frac{\sigma^2_{pre}}{N} + \frac{\sigma^2_{post}} {N^2 \widehat{g}^2}$   \\
		 Average binning &   $\frac{1}{N}$$\sum_{i=1}^{N} \{l_i + n_{i,pre} \}+ \frac{n_{post}}{g}$   $\quad$ & $\frac{l^*}{N} + \frac{\sigma^2_{pre}}{N} + \frac{\sigma^2_{post}} {g^2}$  \\
		Digital binning $\qquad$    &  $\frac{1}{N}$$\sum_{i=1}^{N} \{l_i + n_{i,pre} + \frac{n_{i,post}}{g}\}$    & $\frac{l^*}{N} + \frac{\sigma^2_{pre}}{N} + \frac{\sigma^2_{post}} {N g^2}$\\
\bottomrule
	\end{tabular}
\end{table}

Fig.~\ref{fig:binning-theory}(a) shows one set of signal and noise curves for pitch $p_0$ and at a given light level $l_0$. Considering $\text{SNR}_t=1$, the two curves intersects at the cut-off frequency $f_\text{cutoff}(l_0, p_0)$. This means that given light level $l_0$, a sensor with pixel pitch $p_0$ can resolve features with maximum frequency $f_\text{cutoff}(l_0, p_0)$.
As shown in \cref{fig:binning-theory}(b), we analyze for varying pixel pitches, from \SI{0.5}{\micro\meter} to \SI{2.0}{\micro\meter}, given the same light level $l_0$, and find their cutoff frequency $f_\text{cutoff}(l_0, p)$.
Based on (b), we obtain the pixel and its effective resolution at light level $l_0$, which is shown as the blur curve in (c), and find out the optimal pitch for $l_0$, $p^*(l_0) = \arg\max_p f_\text{cutoff}(l_0, p)$. We repeat this analysis for varying light levels from extremely dark to bright.
As shown in \cref{fig:binning-theory}(d), this allows us to analyze the optimal pixel pitch that achieves the highest effective resolution for each scene light level.
Note that the optimal pixel size is a function of light level, and decreases as the scene gets brighter. This confirms the intuition that small pixels suit bright scenes and large pixels suit dark scenes.
The SNR threshold $\text{SNR}_t$ is a hyperparameter and is determined empirically by examining a set of measured image quality. We use $\text{SNR}_t=4$ for all our experiments.

To evaluate the effectiveness of our binning theory, we simulate a texture patch captured under various light conditions and binning sizes, as shown in \cref{fig:sweep-bin}(a). Under most light conditions, the predicted binning sizes (black boxes) match the one with best image quality (blue box) indicated by LPIPS scores~\cite{zhang2018unreasonable}.

\subsection{A sensor with varying pixel pitches through binning}

We implement the optimal pixel pitches through pixel binning. We discuss three binning types that are commonly seen in sensors --- analog additive, analog average, and digital binning. 

\begin{itemize}[leftmargin=*]
	\item \textit{Analog additive binning.} Both analog binning modes are conducted on the analog signals. After photoelectron counts are converted into analog voltages, the sensor \textit{sum} up the analog voltage of $N$ neighboring pixels. 
	%The combined signal is then amplified by $g$ times by the amplifier and finally read out as digital values.
	
	\item \textit{Analog average binning.} The sensor weighted \textit{average} the analog voltages of $N$ neighbouring pixels.
	\item \textit{Digital binning.} Signals are binned after read out as digital values.
\end{itemize}

We summarize the noise models and total noise variances of all binning modes in \cref{tab:binning-modes}. All three binning modes reduce the photon noise by $N$ times. At first sight, analog additive binning reduces the most read noise, but when we take different gains into account, digital binning becomes the best. This interesting observation comes from the fact that digital binning does not raise the analog voltage while additive binning scales voltage up by around $N$ times. To avoid saturation, additive binning uses a gain that is around $N$ times smaller than that of digital binning. With $\widehat{g} = g/ N$, read noise variance of additive binning becomes $\smash{{\sigma^2_{post}} /({N^2 \widehat{g}^2}) = {\sigma^2_{post}}/ {{g}^2} }$, and is larger than that of digital binning.

Fig.~\ref{fig:sweep-bin}(b) compares analog and digital binning under different gain settings. When gain is restricted to be low, spatially-varying analog binning can significantly increase the SNR for dark patches, where as digital binning suffers to recover details from noise. However, if a larger gain is allowed, digital binning is superior to analog binning as it reduce noise without sacrificing resolution.

\section{Emulated Results on Real Hardware}
\label{sec:simulated}

\begin{figure}[t]
	\includegraphics[width=1.0\linewidth]{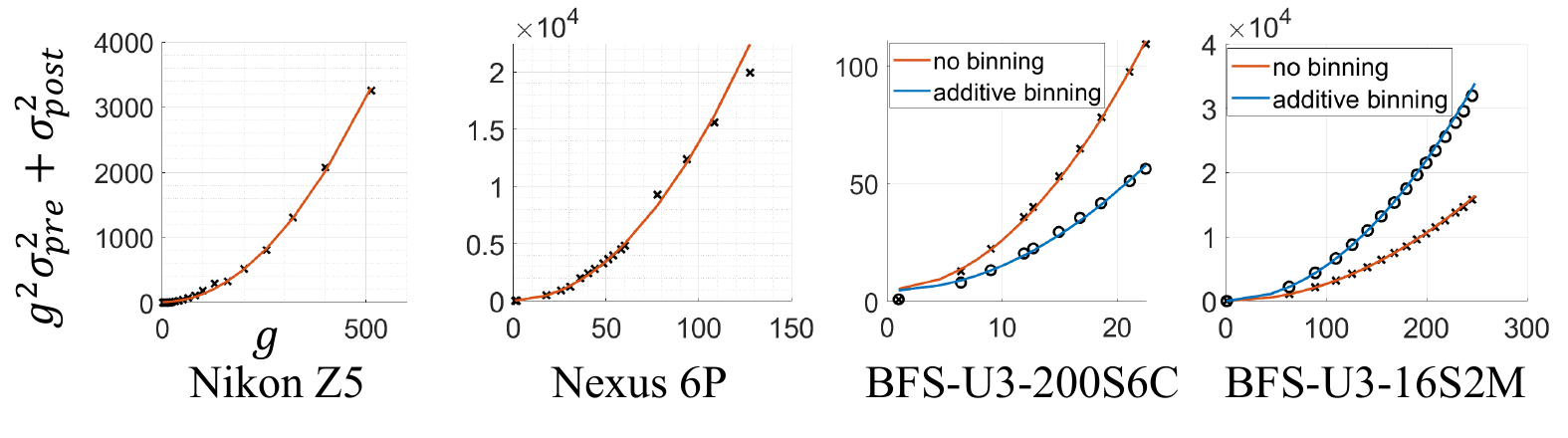} 
	\caption{\textbf{Pre- and post-amplifier read noise calibration.} $x$-axis is gain and $y$-axis is the total noise variance. Base ISOs are normalized to 1.}
	\label{fig:sensor-calib}
\end{figure}

\begin{table}[t]
	\footnotesize
	\caption{\textbf{A summary of sensors.} $\sigma_{pre}$ and $\sigma_{post}$ are in analog digit unit.}
	\centering
	\ra{1.2}
	\begin{tabular}{@{}llllcc@{}} \toprule
		Camera & Type & Max. Gain & Binning & $\sigma_{pre}$& $\sigma_{post}$ \\
		\midrule
		Nikon Z5 & Mirrorless & ISO 51200 $\ \ $ & --- & 0.11 & 3.53 \\
		Nexus 6P & Smartphone & ISO 7656 & --- & 1.17 & 7.39 \\
		\multirow{2}{*}{BFS-U3-200S6C} $\ $ & \multirow{2}{*}{Machine Vision}  $\ $ & \multirow{2}{*}{27 dB} & $1\times 1$ & 0.46 & 2.27 \\
		& & & $2\times 2$ (avg) & 0.32 & 2.17 \\
		\multirow{2}{*}{BFS-U3-16S2M} & \multirow{2}{*}{Machine Vision} & \multirow{2}{*}{48 dB} & $1\times 1$ & 0.52 & 4.55 \\
		%& & & $2\times 2$ (add) $\ $ & 0.74 & 4.73 \\
		& & & $2\times 2$ (add) $\ $ & 0.23 & 1.47 \\
		\bottomrule	
	\end{tabular}
	\label{tab:sensor-calibration}
\end{table}

\paragraph{Sensor Calibration.} 
We calibrate and capture data with four cameras: Nikon Z5 (mirrorless camera), Nexus 6P (smartphone main camera), FLIR BFS-U3-200S6C (machine vision  with Bayer color filter arrays), and FLIR BFS-U3-16S2M (machine vision monochrome). 
Table~\ref{tab:sensor-calibration} summarizes their key specifications.
We calibrate their pre- and post-amplifier read noise by closing the cap on sensors and capturing dark noisy frames, 
and capturing images with varying gain.
As shown in \cref{fig:sensor-calib}, we fit a quadratic curve between $\sigma^2$ and $g$ and estimate the coefficients $\sigma_{pre}$ and $\sigma_{post}$.
As the two machine vision cameras also support analog binning, we repeat the calibration with analog binning.
We summarize the calibrated noise statistics in Table~\ref{tab:sensor-calibration}. 
We can see that post-amplifier read noise is usually around one magnitude larger than pre-amplifier noise.

\begin{figure*}[t]
	\centering
	\includegraphics[width=1.0\linewidth]{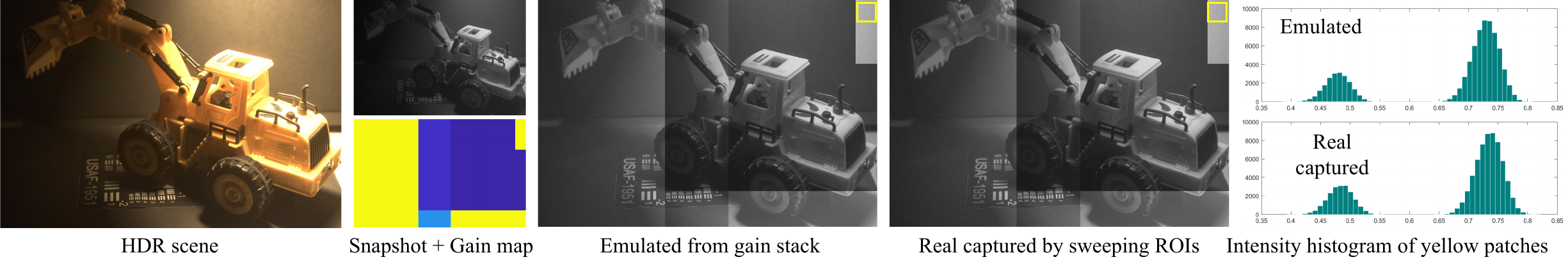}
	\caption{\textbf{Emulator versus real-capture by windowing.} All images are captured by BFS-U3-200S6C camera. The emulated image is composed from a real captured gain stack. Real capture is obtained by sequentially setting ROIs using low-level API, integrating, and read out with the optimal gain.}
	\label{fig:scan}
\end{figure*}

\begin{figure*}[t]
	\centering
	\includegraphics[width=0.63\linewidth]{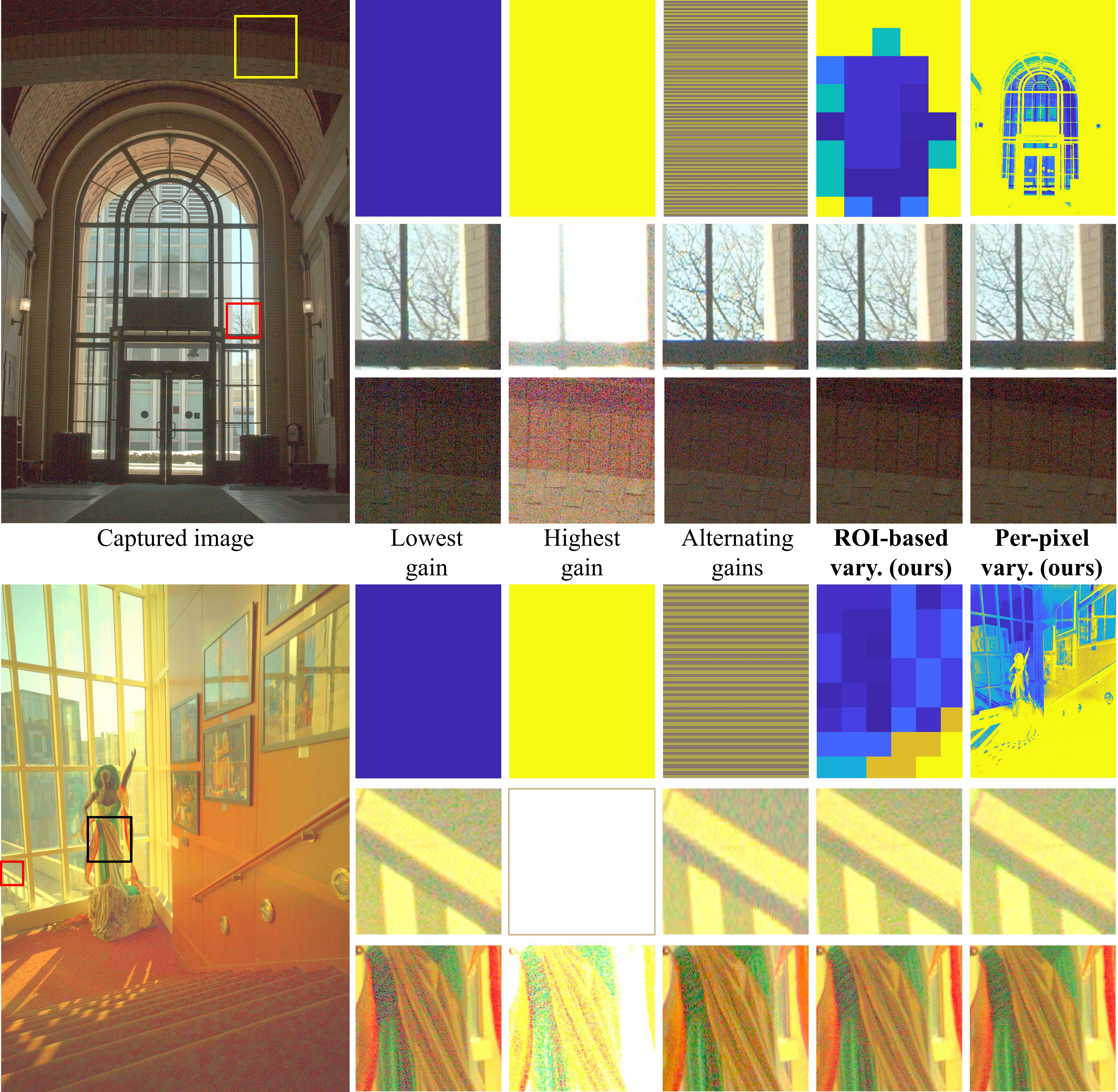}
	\caption{\textbf{Comparisons of gain modes in HDR.} Upper figures are captured by Nikon Z5 and lower by Nexus 6P. The lowest and highest ISOs for Nikon Z5 are 100 and 51200 and for Nexus 6P are 60 and 7656.}
	\label{fig:sv-gain}
\end{figure*}

\begin{figure*}[t]
	\includegraphics[width=1.0\linewidth]{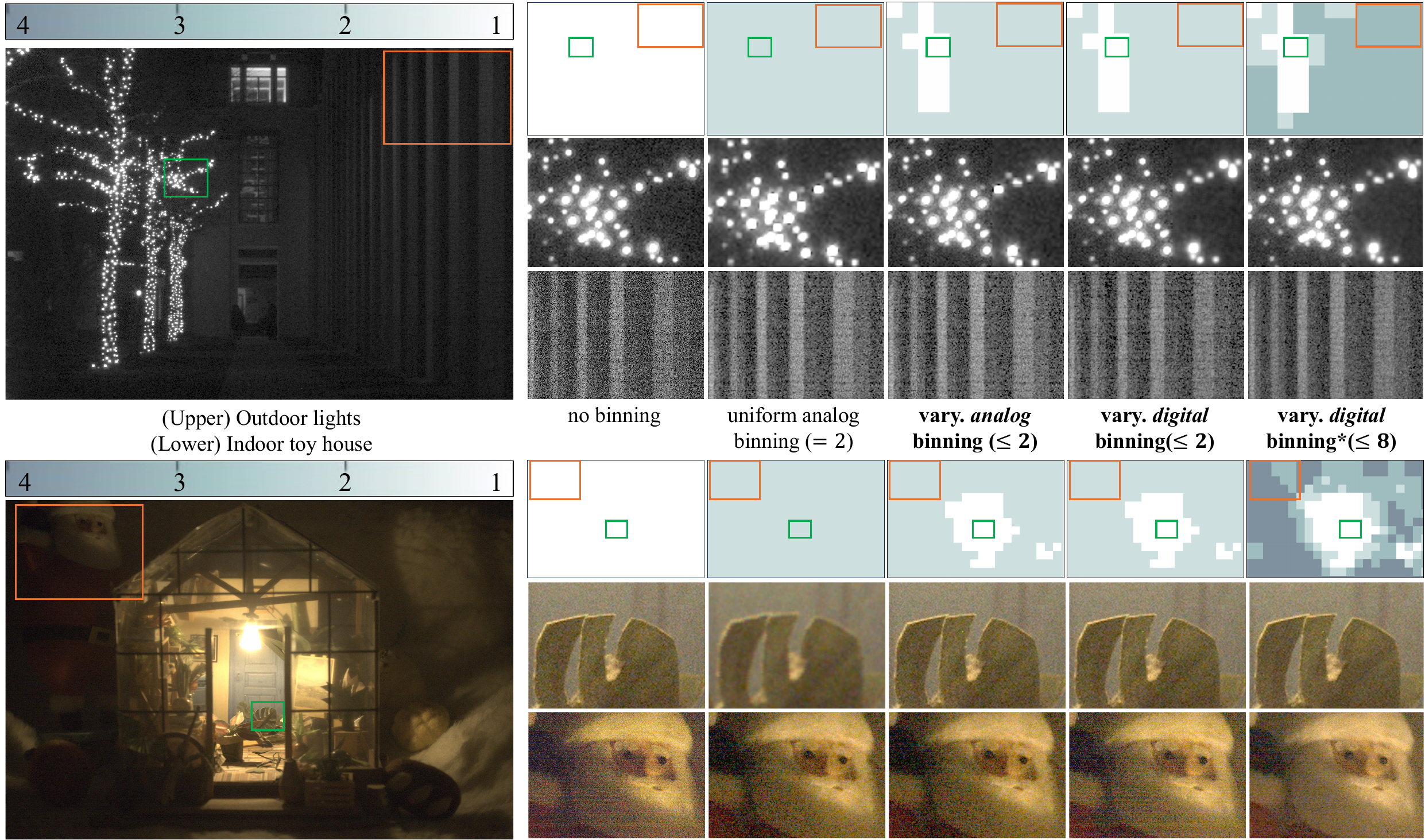} 
	\caption{\textbf{Comparisons of binning modes in HDR.} Upper is captured by BFS-U3-16S2M and lower BFS-U3-200S6C. The first two columns are from off-the-shelf binning modes and the last three columns are proposed spatially-varying binning. Since both cameras only support analog binning up to $2\times 2$, we demonstrate binning with larger sizes on digital binning. *The last column is digital binning under varying gain.
	}
	\label{fig:sv-bin}
\end{figure*}

\paragraph{Emulator.} 
We emulate spatially-varying gain and binning by capturing a gain stack, sweeping from the lowest to highest gain settings, computing gain maps and binning maps based on the lowest gain snapshots, and compositing ROIs from corresponding bursts into one image.
For mirrorless and smartphone cameras, we only capture one gain stack, and for machine vision cameras, we capture  gain stacks with and without analog binning.

Note that both machine vision cameras supports random access to region-of-interests (ROIs), and allow user to specify gain and binning for each ROI. We can thus sequentially set ROI and readout each ROI using optimized parameters. In \cref{fig:scan}, we show that for ROI-based spatially-varying techniques, emulating from a gain stack produce the same noise statistics as real-capture through windowing.

\paragraph{Effect of spatially-varying gain.} 
Shown in \cref{fig:sv-gain}, we compare constant gain, alternating gain~\cite{hajsharif2014hdr}, and the proposed spatially-varying gain on Nikon Z5 and Nexus 6P. 
%Camera settings except for ISO are the same for all methods.
For each scene and from top to bottom, we show gain map, one bright patch, and one dark patch.
When captured with lowest ISO, dark regions are excessively noisy and the details are largely degraded by noise. %For example, the bricks in the upper set and the cloth's wrinkles in the lower set are hard to tell.
When captured with highest ISO, read noise is suppressed, but the bright regions are over-exposed.
Alternating ISOs emulates a sensor with lowest and highest ISOs every alternating rows~\cite{hajsharif2014hdr}.
However, their vertical resolution is reduced by half, since only half rows are valid for the bright regions and extremely dark regions.
Finally, the proposed spatially-varying gain captures details in the bright region without sacrificing resolution, and effectively reduce sensor noise in the dark regions.
This reflects an expansion of the sensor dynamic range without reducing resolution.

\paragraph{Effect of spatially-varying binning.} 
We evaluate the effect of spatially varying binning on BFS-U3-16S2M monochrome camera and BFS-U3-200S6C color camera.
As shown in the first two columns in \cref{fig:sv-bin}, we capture HDR scenes with and without $2 \times 2$ analog binning.
Without binning, regions in the dark are extremely noisy. With uniform analog binning, the noise is reduced, but it sacrifices the resolution in the bright regions.
For example, point lights appear square-ish in the upper example, and the leaves appear blurry in the lower example.
Compared with those without binning, our methods reduce noise in the dark regions, and compared to those from uniform binning, ours retain appealing details in the bright regions.
With the same gain, digital binning is slightly worse than analog additive binning, which confirms our analysis.
Finally, we examine spatially-varying binning up to size eight, using digital binning since analog binning only supports up to size two; this 
achieves the best performance. This suggests that when light conditions is extremely low, image sensors could benefit from analog additive binning larger than two. If large gain is allowed, conducting digital binning would produce the best quality.

\begin{figure*}[t]
	\centering
	\includegraphics[width=\linewidth]{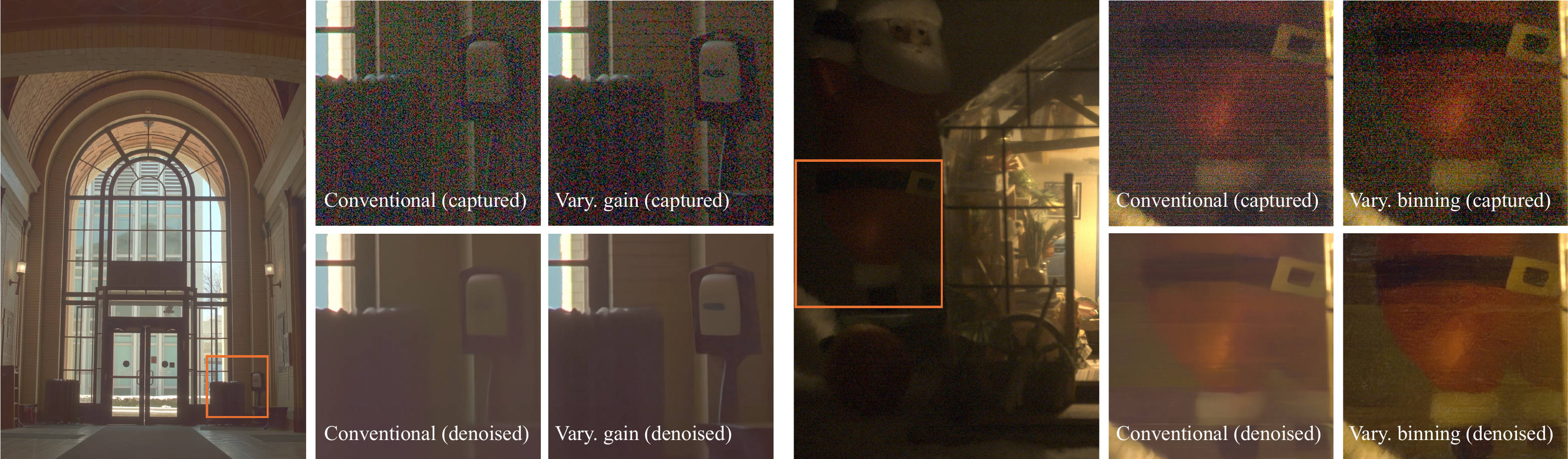}
	\caption{\textbf{Effect of transformer-based restoration networks.} Left example is captured by Nikon Z5 and right BFS-U3-200S6C. The upper row shows the captured image after demosaicking and gamma correction, and the lower row shows the above images denoised by Restormer~\cite{zamir2022restormer}. (Left) Compared to conventional sensor, the proposed spatially-varying gain recovers much more object details in the dark lighting; (Right) Compared to conventional, spatially-varying binning retains better contrast and recovers sharper contours.}
	\label{fig:restormer}
\end{figure*}

\paragraph{Effect of post-processing.} Fig.~\ref{fig:restormer} compares images denoised by Restormer, a SOTA transformer-based image restoration network~\cite{zamir2022restormer}. We use the pretrained "Real denoising" checkpoint and denoise captured images with a patch size of $720\times 720$ and an overlap of $32$ pixels. The captured images are demosaicked and gamma corrected before fed into the network. For spatially-varying binning (up to $2\times 2$ analog binning) images, we feed both the full resolution and downsampled images to the network, so that regions without binning and with binning are denoised separately, and then merge two denoised copies according to the binning map.

\paragraph{Teaser.}  As shown in Figure~\ref{fig:new-teaser}, we capture with an BFS-U3-200S6C with no binning and constant gain, with proposed varying gain, the proposed varying analog additive binning, and denoise all by Restormer~\cite{zamir2022restormer}.
The baseline image is extremely noisy, leaving the alphabets hard to tell and hazing colors in the cropped patch.
Adapting gain to local brightness can effectively reduce noise and we can the details clearer.
Applying varying binning reduces noise and improves the signal contrast compared to baseline.

\begin{figure*}[t]
	\centering
	\includegraphics[width=0.9\linewidth]{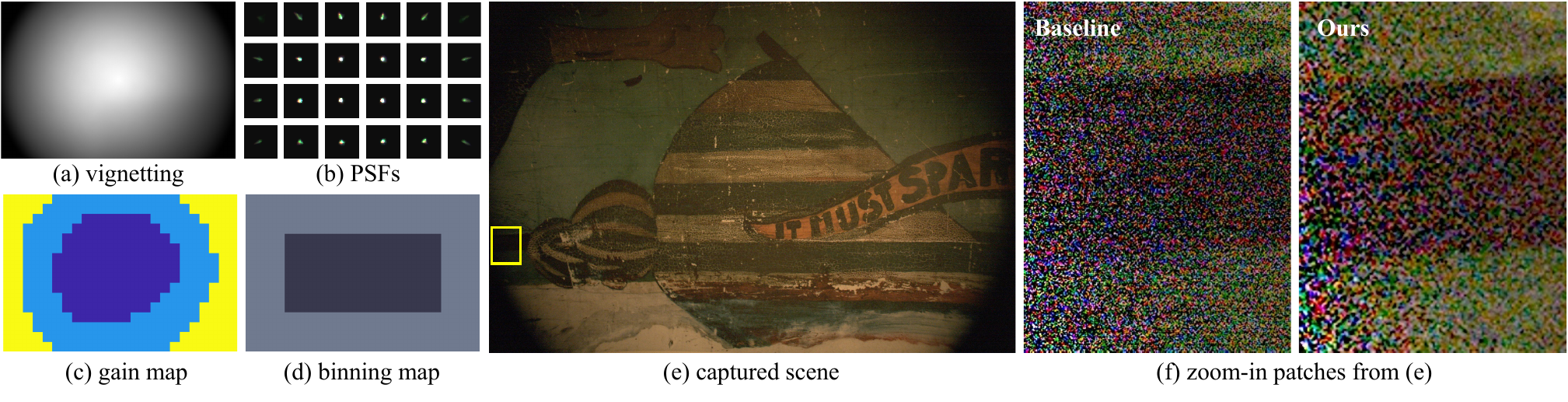}
	\vspace{-0.8em}
	\caption{\textbf{Proposed techniques for vignetting and lens blur.} Images are captured by BFS-U3-200S6C camera with an \SI{8}{\milli\meter} C-mount lens and $f/1.4$ aperture. From zoom-in patches (f), we can see that the proposed spatially-varying gain and binning reduces the extend of noise and produces better contrast compared to the baseline.}
	\label{fig:real-vignette}
\end{figure*}

\paragraph{Application on vignetting and lens blur.} Vignetting and spatially-varying blur is common in photography. 
We show that the proposed spatially-varying gain and binning can increase the noise performance for vignetting and lens blur. 
As shown in \cref{fig:real-vignette}(a)(b), we pre-calibrated the vignetting map and spatially-varying lens blur for BFS-U3-200S6C with an \SI{8}{\milli\meter} lens and fixed aperture $f/1.4$.
We compute a gain map that is inversely proportional to vignetting and a binning map that bin $2\times2$ pixels when the PSFs are flat.
We capture a scene shown in (e) and correct the intensity by inverting the vignetting map.
(f) shows zoom-patches for baseline no binning and constant gain and proposed spatially-varying gain and binning. We can see that with the proposed technique, noise is reduced to a much smaller region towards the edge.

\section{Quantitative Results}
\label{sec:quantitative}
\begin{table}[t]
	\footnotesize
	\caption{\textbf{Quantitative results on simulated HDR scenes~\cite{fairchild2008}.} For each method and each scene, we show worst-case  SSIM (larger is better) and worst-case LPIPS~\cite{zhang2018unreasonable} scores (lower is better).}
	\centering
	\ra{1.2}
	\begin{tabular}{@{}lllll@{}}
		\toprule
		\multirow{2}{*}{Scenes} & const gain & \textbf{vary. gain} & const gain & \textbf{vary. gain}\\
		& no bin. & no bin. & \textbf{vary. bin} & \textbf{vary. bin.}\\
		\midrule
		BarHarborSunrise.exr & 0.03/ 1.47 $\ $ & 0.07/ 1.29 $\ $ & 0.14/ 1.14 $\ $ & \gbox{0.24}/ \rbox{0.97} \\ 
		BloomingGorse.exr & 0.47/ 0.46  & 0.62/ \rbox{0.37}  & 0.47/ 0.46  & \gbox{0.63}/ \rbox{0.37} \\ 
		GoldenGate.exr & 0.04/ 1.39  & 0.07/ 1.22  & 0.07/ 1.08  & \gbox{0.09}/ \rbox{1.00} \\ 
		JesseBrownsCabin.exr $\ $ & 0.04/ 1.40  & 0.08/ 1.20  & 0.08/ 1.25  & \gbox{0.09}/ \rbox{1.12} \\ 
		MirrorLake.exr & 0.20/ 0.93  & 0.48/ 0.56  & 0.38/ 0.66  & \gbox{0.57}/ \rbox{0.50} \\ 
		NiagaraFalls.exr & 0.51/ 0.60  & 0.74/ \rbox{0.40}  & 0.68/ 0.45  & \gbox{0.80}/ 0.41 \\ 
		RedwoodSunset.exr & 0.03/ 1.38  & 0.08/ 1.22  & 0.23/ 0.81  & \gbox{0.26}/ \rbox{0.78} \\ 
		TunnelView.exr & 0.40/ 0.57  &  \gbox{0.61}/ 0.38  & 0.50/ 0.47  & \gbox{0.61}/ \rbox{0.36} \\ 
		WallDrug.exr & 0.03/ 1.35  & 0.10/ 1.12  & 0.12/ 1.14  & \gbox{0.32}/ \rbox{0.82} \\ 
		YosemiteFalls.exr & 0.06/ 1.24  & 0.16/ 1.01  & 0.10/ 1.04  & \gbox{0.33}/ \rbox{0.72} \\ 
		\midrule
		Average & 0.18/ 1.08 & 0.30/ 0.88 & 0.28/ 0.85 &  \gbox{0.39}/ \rbox{0.71}\\
		\bottomrule	
	\end{tabular}
	
	\label{tab:quantitative-lpips}
\end{table}

In \cref{tab:quantitative-lpips}, we compare conventional sensor with the proposed spatially-varying gain, spatially-varying binning, and the combined. Quantitative metrics are computed on simulated images. We download high-quality HDR images from ~\citet{fairchild2008} and simulate noisy captured images based on the noise model in \cref{eq:noise-model}. 
%Please refer to the appendix for the specifications of the simulator.
The simulated camera has a pixel pitch of \SI{0.5}{\micro\meter}, full well capacity of 1000 $e^-$, pre- and post-amplifier read noise of $\sigma_{pre} = 0.33 e^-, \sigma_{post} = 3.31e^-$. We set the black level to be $5\%$ and read out $12$-bit images. We normalize the captured image such that the mean intensity equals to $0.05$, and gamma correct the normalized images with a power of $1/3.2$.
 We use an ROI size of $128 \times 128$. $\text{SNR}_t=4$ is used to decide the optimal binning.  We compare the gamma corrected captured images with the ground-truth ones, and compute LPIPS~\cite{zhang2018unreasonable} and SSIM. We compute the metrics for each ROI and take the worst-case performance. Numbers to the left of slash is SSIM and right is LPIPS. LPIPS is smaller better and  SSIM is larger better. We highlight the best SSIM and LPIPS using green and red box for each scene. We see that the proposed spatially-varying gain and binning is significantly better than conventional sensors.

\section{Discussion}
\begin{figure}
	\centering
	\includegraphics[width=\linewidth]{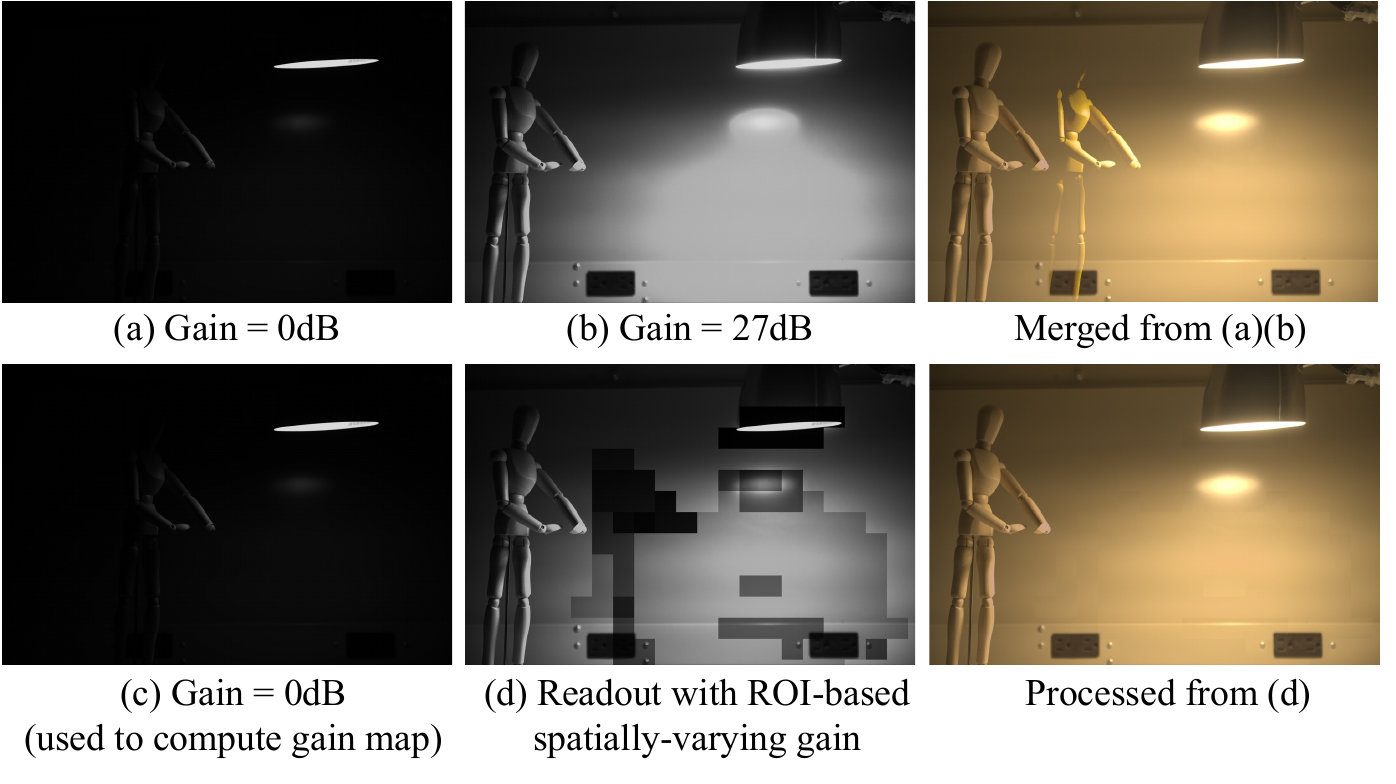}
	\caption{\textbf{Comparison between multi-shot technique and ours.} (Upper) Multi-shot method tends to produce ghosting artifacts with the presence of dynamic objects.
	(Lower) The proposed spatially-varying gain technique utilizes the first capture to determine gain map. The final capture is soley processed from frame (d) by normalizing gains.}
	\label{fig:motion}
\end{figure}
We propose two novel readout techniques for  image sensors: spatially-varying gain and binning that adapt to the local scene brightness. The proposed techniques significantly improve the noise performance of the captured  images for low-light regions, thereby expanding the sensor dynamic range.

\paragraph{Comparison with multi-shot techniques.}
There is a rich literature that captures high-quality HDR scenes through exposure or gain bracketing~\cite{perez2022ntire,hasinoff2010noise}. However, multi-shot techniques are sensitive to motion, and aligning dynamic objects across frames requires significant computation in post-processing. In contrast, our spatially-varying techniques are more robust for dynamic scenes. An example is shown in \cref{fig:motion}. For a fair comparison, we capture two frames for both methods. In the upper row, the multi-shot technique merges low- and high-gain frames through post-processing, and is prone to produce ghosting artifacts. In contrast, our technique (lower row) uses the first frame only to compute the gain map and guide the readout of the subsequent frame. Thereby, ours is free of ghosting.

\paragraph{What does it take to implement in hardware?}
First, the proposed ROI-based techniques are partially implementable using off-the-shelf CMOS sensors~\cite{flir} through windowing, as shown in \cref{fig:scan}. 
However, existing sensors reset the cycle of integration after each ROI readout, which could lead to motion artifacts for dynamic scenes.
To avoid exposing the sensor repeatly during readout, the internal timing should restart integration only after all ROIs are sequentially read out.
Second, implementing the proposed per-pixel varying gain requires more engineering efforts as it requires a rapid variable amplifier and additional circuitry to set gain based on previous readout. Previous works~\cite{cui2017high,lee2007wideband} demonstrates ultra-wideband programmable variable gain amplifier. They are controled by input signal and can reach a bandwith up to $900$MHz, offering a promising solution.

\paragraph{Is analog binning really superior to digital?}
Prior works emphasize the advantage of analog binning over digital binning. Our analysis shows that this conclusion is arguable with the interplay of gain. When the scene is dark and gain is restricted to a small value, analog binning is indeed better than digital binning, by combining the signal levels to overcome read noise. However, when a larger gain is allowed, either by the proposed spatially-varying gain or other dual ISO techniques, digital binning is all you need to improve the noise performance.
\label{sec:discuss}

\section*{Acknowledgements} This work was supported by Global Research Outreach program of Samsung Advanced Institute of Technology and the NSF CAREER award CCF-1652569.

% Bibliography
\bibliographystyle{ACM-Reference-Format}
\bibliography{references}

\cleardoublepage
\appendix
\section{Appendix}
\subsection{Detailed derivation of optimal pixel pitch}

We characterize the scene contents of various feature sizes by examining sinusoidal signals with varying frequencies.
Consider a scene of sinusoidal function that varies at frequency $f_0$ cycles/\SI{}{\milli\meter} in $x-$direction and constant in $y$-direction and has the maximum intensity of $l_0 e^-/$\SI{}{\milli\meter^2} within the exposure time.
$$i^*(x,y;f_0,l_0) = \frac{l_0}{2}\cos(2\pi f_0 x) + \frac{l_0}{2}$$
Consider a camera with an ideal lens and a sensor with a pixel pitch \SI[parse-numbers=false]{p}{\micro\meter}. The measured signal can be modeled as a convolution between the signal and a box function induced by the pixel size,
\begin{equation}
	l(x,y;f_0, l_0, p) = i^*(x,y;f_0,l_0) * b \biggl\{ \frac{x}{p}, \frac{y}{p}\biggr\} + n_{shot}(x,y) + n_{read}(x,y).
\end{equation}
The first part models the expected noise-free signal measurement by incorporating the blurring effect of pixel pitch.
We obtain the expression of the measured signal by taking the Fourier transform of $i^*$ and $b$, multiplying them, and taking the inverse Fourier transform. The expression for noise-free component is,
$$
l^*(x,y;f_0,l_0) = \frac{l_0p}{2} \frac{\sin(\pi p f_0)}{\pi f_0} \cos(2\pi f_0 x) + \frac{l_0}{2}p^2
$$
As $-1\leq \cos(2\pi f_0 x) \leq 1$, $l^*$ has a maximum and minimum intensity of,
\begin{align*}
	& \max l^* = \frac{l_0p}{2} \frac{\sin(\pi p f_0)}{\pi f_0} + \frac{l_0}{2}p^2 \\
	& \min l^* = -\frac{l_0p}{2} \frac{\sin(\pi p f_0)}{\pi f_0} + \frac{l_0}{2}p^2
\end{align*}
The contrast of the noise-free signal is
\begin{equation}
	c(f_0; l_0, p) = |\max l^* - \min l^*| = l_0 p^2 \frac{sin(\pi p f_0)}{\pi f_0}.
\end{equation}
Next, we examine the total noise variance. Note that we approximate the Poisson distributed shot noise by Gaussian distribution $n_{shot}(x_0,y_0) \sim \mathcal{N}(0, l^*(x_0, y_0))$ with a mean zero and variance of latent signal.
Interestingly, $n_{shot}(x_0,y_0), n_{shot}(x_0+T_0,y_0), ..., n_{shot}(x_0+NT_0,y_0), x_0\in[0,T_0), N\in\mathbb{Z}$ can be viewed as iid samples of the same distribution $\mathcal{N}(0, l^*(x_0, y_0))$, as the latent signal $l^*$ is a periodic function with period $T_0$, $l^*(x_0, y_0) = l^*(x_0 + T_0, y_0) = ... = l^*(x_0+NT_0, y_0), N \in \mathbb{Z}$.
Therefore, the \textit{combined} variance of $n_{shot}(x,y_0) \forall x\in [0, T_0)$ can be written as,
\begin{align*}
	\sigma_{shot}^2 &= \int_{x=0}^{T_0} Var(n_{shot}(x,y_0)) \frac{1}{T_0} dx \\
	&= \int_{x=0}^{T_0} l^*(x,y_0) \frac{1}{T_0} dx \\
	&= \int_{x=0}^{T_0} \Bigg(\frac{l_0 p}{2} \frac{\sin(\pi p f_0)}{\pi f_0} \cos(2\pi f_0 x) + \frac{l_0}{2}p^2\Bigg) \frac{1}{T_0} dx \\
	& = \frac{l_0}{2} p^2.
\end{align*}
$n_{read}$ includes both pre- and post-amplifier read noise and follows a Gaussian distribution $n_{read} \sim \mathcal{N}(0, \sigma^2_{pre} + \sigma^2_{post} / g^2)$. Putting the combined variance of shot and read noise, we obtain the total noise variance as the following,
\begin{equation}
	\sigma^2(f_0; l_0, p)= {\sigma^2_{pre} + \sigma^2_{post} / g^2 + l_0 p^2 / 2},
\end{equation}
where read noise is independent of pixel size, and shot noise variance increase proportionally to pixel area $p^2$. Interestingly, the combined noise variance is independent of the underlying signal frequency $f_0$.

With the noise-free signal contrast $c(\cdot)$ and the combined noise standard deviation $\sigma(\cdot)$, we can find out the cutoff frequency that has an $\text{SNR}_t$ for a given pixel pitch $p$ and at light level $l_0$,
\begin{equation}
	f_\text{cutoff}(l_0, p)= \arg\min_{f_0} \Biggl\| \frac{c(f_0; l_0, p)}{\sigma(f_0; l_0, p)} - \text{SNR}_t \Biggr\|.
\end{equation}

\subsection{Details about tonemapping used in the paper}

As shown in~\cref{tab:tonemap}, we list the tonemapping functions used in each figure.
~\cref{tab:tonemap} lists the tonemapping functions used in different figure for the cameras whose properties are summarized in ~\cref{tab:camera-specs} and ~\cref{tab:sensor-calibration}. For Nikon Z5 and Nexus 6P, we use built-in Matlab function \texttt{tonemapfarbman}~\cite{farbman2008edge} with \texttt{RangeCompression=0.2, Saturation=2.5}. For machine vision cameras, we use power functions. In zoomed-in patches, we magnify the intensities of dark patches by $500$ times so that the noise and details in the dark regions are more visible, and we only magnify bright patches by $50$ to avoid saturation.

\begin{table}
	\caption{\textbf{Tonemapping functions used for different cameras/
results.}}
	\centering
	\ra{1.2}
	\begin{tabular}{@{}lllll@{}}
		\toprule
		Camera & Tonemap & Figures \\
		\midrule
		Nikon Z5 & \citet{farbman2008edge} & Fig.~\ref{fig:sv-gain} upper (2-3 row) \\
		Nexus 6P & \citet{farbman2008edge} & Fig.~\ref{fig:sv-gain} lower (2-3 row) \\
		\multirow{2}{*}{BFS-U3-200S6C} $\ $ & $(50 \cdot i)^{1/2.2}$ & Fig.~\ref{fig:sv-bin} upper (2nd row) \\
		& $(500 \cdot i)^{1/2.2}$ & Fig.~\ref{fig:sv-bin} upper (3rd row) \\
		\multirow{2}{*}{BFS-U3-16S2M} & $(50 \cdot i)^{1/2.2}$ & Fig.~\ref{fig:sv-bin} lower (2nd row) \\
		& $(500 \cdot i)^{1/2.2}$ $\ $ & Fig.~\ref{fig:sv-bin} lower (3rd row)\\
		\bottomrule	
	\end{tabular}
	\label{tab:tonemap}
\end{table}

\begin{table}
	\caption{\textbf{Camera specifications.}}
	\centering
	\ra{1.2}
	\begin{tabular}{@{}lllll@{}}
		\toprule
		Camera & Resolution & ROI size \\
		\midrule
		Nikon Z5 & $2640 \times 3960$ & $512 \times 512$ $\ \ $ \\
		Nexus 6P & $3024 \times 4032$ & $512 \times 512$ $\ \ $ \\
		BFS-U3-200S6C $\ $ & $3648 \times 5472$  $\ $ & $256 \times 256$ \\
		BFS-U3-16S2M & $1080 \times 1440$ & $128 \times 128$ \\
		\bottomrule	
	\end{tabular}
	\label{tab:camera-specs}
\end{table}

\end{document}